\documentclass[runningheads]{llncs}
\usepackage[T1]{fontenc}
\usepackage{graphicx,verbatim}
\usepackage{booktabs}
\usepackage{amsmath}

\begin{document}

\title{Intersectional Disentangling of Temporal and Acquisition Bias in Fetal Ultrasound}

\author{
Aya Elgebaly\inst{1} \and
Joris Fournel\inst{1} \and
Benjamin Laine J{\o}nch Jurgensen\inst{1} \and
Kamil Mikolaj\inst{1} \and
Anders Christensen\inst{1} \and
Martin Tolsgaard\inst{2} \and
Claes Ladefoged\inst{1} \and
Aasa Feragen\inst{1}
}

\authorrunning{Elgebaly et al.}

\institute{
Technical University of Denmark, Denmark \and
CAMES Rigshospitalet, Denmark \\
\email{
aafel@dtu.dk,
afhar@dtu.dk
}
}
\maketitle

\begin{abstract}
Fairness studies of medical imaging AI often explain subgroup performance gaps through under-representation in the training data. We show that intersectional analysis can disentangle fairness and performance gaps arising from clinical and acquisition confounders that co-vary with the target. As a case, we study scan-time fetal weight estimation from obstetric ultrasound, analyzing two models: a state-of-the-art deep learning (DL) model and the clinical gold-standard Hadlock formula. Using unsupervised slice discovery, we find that high-error subgroups share extreme in the image-acquisition pixel spacing (PS) and in the scan-to-delivery (STD) interval. Of these, PS is an acquisition parameter that can be optimized, while STD is a potential confounder for both PS and our bias diagnostics. Subgroup inspection alone cannot separate them. We disentangle the factors using a model-agnostic analysis with identical metadata partitions and partial regression. Holding STD fixed, the apparent PS effect collapses to a small residual (standardized coefficient $\beta=-0.17$), whereas holding PS fixed, STD dominates error ($\beta=+0.56$). Both models degrade with increasing STD, including the biometric formula, indicating much of the error is intrinsic to the prediction target rather than imaging. The DL model is $\sim$1.5$\times$ more sensitive to STD than Hadlock, though it remains more accurate in every subgroup. We conclude that fairness analyses need to carefully analyze potential confounds, or risk attributing an effect such as temporal or acquisition-related dependency to demographics.
\end{abstract}

\begin{figure}[t]
\centering
\includegraphics[width=\textwidth]{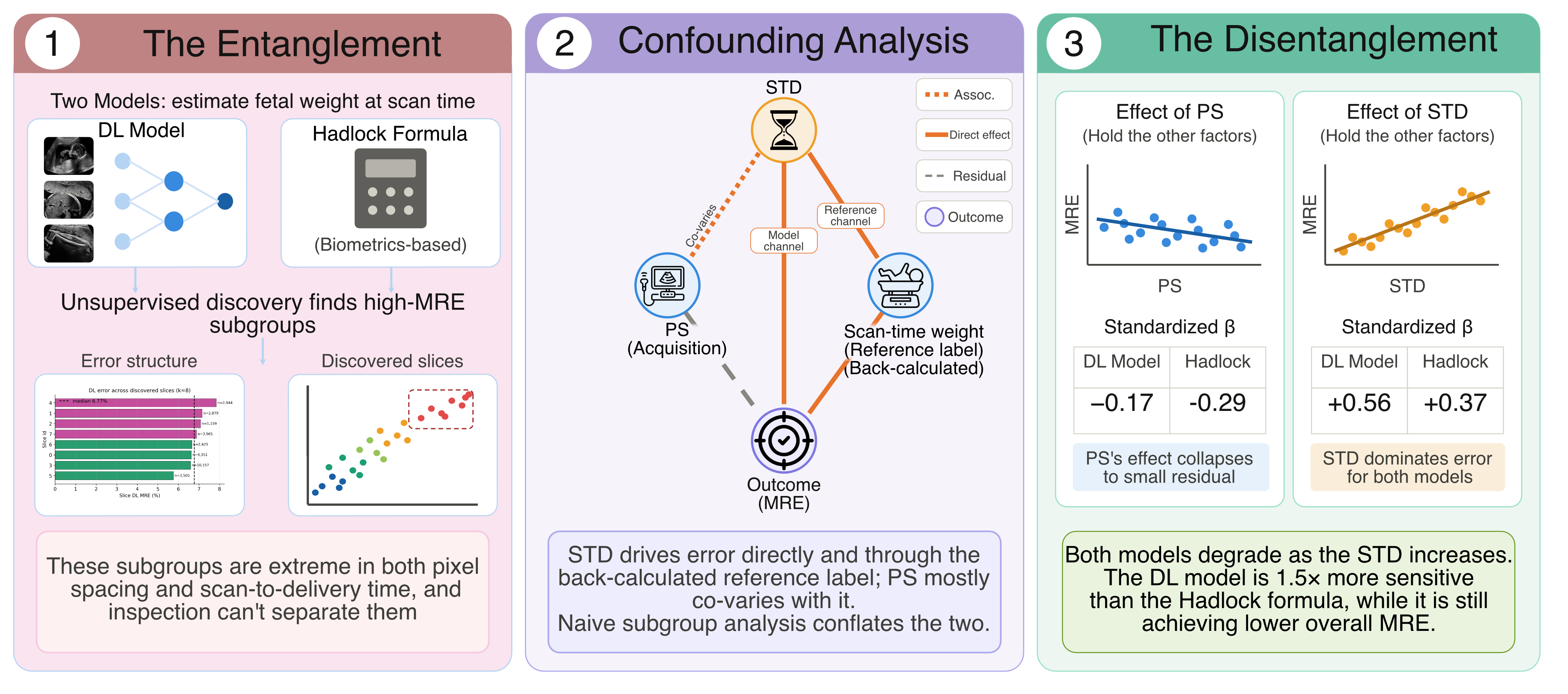}
\caption{Overview. (1) Two estimators predict scan-time fetal weight; unsupervised slice discovery surfaces high-error subgroups extreme in \emph{both} pixel spacing (PS) and the scan-to-delivery (STD) interval. (2) STD is clinically driven and co-varies with PS, confounding naive subgroup analysis. (3) After adjustment, PS collapses to a small residual while STD dominates error for both models, with the deep learning model $\sim$1.5$\times$ more sensitive to the interval though more accurate overall.}
\label{fig:teaser}
\end{figure}

\section{Introduction}
Ultrasound estimation of fetal weight is central to detecting abnormal growth, where small- or large-for-gestational-age status carries substantial perinatal risk ~\cite{andreasen2021we,hugh2021reduction}. Deep learning (DL) now exceeds the long-standing Hadlock biometry formula at this task~\cite{mikolaj2025predicting}, intensifying the question of whether learned models are fair across patients and imaging conditions.

It is very common for fairness work in medical imaging to explain subgroup performance gaps through \emph{data representation}: groups under-represented at training are predicted less accurately~\cite{larrazabal2020gender,seyyed2021underdiagnosis,roski2019artificial,nazer2023bias,norori2021addressing}, and dataset balance is often highlighted as a mitigation strategy. Yet, we know that representation imbalance is not the only mechanism: disparities can persist after balancing~\cite{dawood2025racial,glocker2023risk}, and unsupervised ``slice discovery'' methods routinely surface high-error subgroups whose defining factor is not a demographic, but a technical or clinical variable~\cite{eyuboglu2022domino,oakden2020hidden,olesen2024slicing,johnson2023does}. A subgroup gap may therefore reflect a \emph{confound}: a clinically driven variable that co-varies with the factor under suspicion, obstructing our interpretation of performance disparities.

We make this concrete in fetal weight estimation. On the one hand, acquisition conditions vary through the pregnancy, and sonographers enlarge the field of view, raising pixel spacing (PS, the physical distance each pixel represents), for larger and later fetuses, creating a correlation between acquisition parameters and fetal weight. At the same time, since the scan-time weight reference label is not actually available, it is constructed by back-calculating the recorded birth weight along a population growth curve~\cite{marvsal1996intrauterine}. As a result, the further a scan precedes delivery, the more this reference must bridge, and the more individual growth may deviate from the population average. The scan-to-delivery (STD) interval, the number of days between the scan and delivery, thus becomes a candidate driver of measured error. Finally, the acquisition parameters and STD co-vary, making it hard to assess the origins of performance differences through a fairness audit that examines demographic subgroups in isolation.

Our contributions are (summarised in Figure~\ref{fig:teaser}):
\begin{itemize}
\item \textbf{Entanglement.} We show that unsupervised bias discovery surfaces high-error subgroups that are jointly acquisition- and time-extreme, so inspection of individual factors alone cannot separate the responsible factor.
\item \textbf{Disentanglement via intersectional analysis.} We combine intersectional partitioning with partial regression, to disentangle temporal and acquisitional factors. In our case study, this concludes that STD, not PS, dominates error, with PS reduced to a small but real residual after adjustment.
\item \textbf{Model-agnostic analysis.} Our analysis is applied identically to DL model and to the biometry-based Hadlock formula. By comparing these, we argue that much of the temporal effect is intrinsic to the prediction target, while a model-specific component is amplified by DL.
\end{itemize}

\section{Materials and Methods}

\subsubsection{Data.}
We analyse the held-out test set of a population-based fetal growth study comprising third-trimester ultrasound examinations from 17 hospitals acquired over a decade~\cite{mikolaj2025predicting}. The unit of analysis is a biometric triplet: the head, abdomen, and femur planes that together constitute one model input. We study $N=31{,}386$ complete triplets ($\geq 28$ gestational weeks), comprising $\sim$94{,}000 ultrasound images, drawn from $6{,}513$ pregnancies of $6{,}220$ women, as some are scanned more than once. For each triplet, the DL model takes the three images as input, whereas the Hadlock formula takes the standard biometric measurements (head circumference, abdominal circumference, femur length) measured manually by clinicians with on-screen calipers during the examination~\cite{hadlock1985estimation}. We additionally record gestational age (GA), maternal body-mass index (BMI; recorded at the examination, with zero values excluded as missing), maternal age, parity, and an averaged PS computed as the geometric mean of per-plane axial and lateral spacing. The reference weight at scan time is obtained by projecting the observed birth weight back to the scan GA along the Mar\v{s}\'al fetal growth curve~\cite{marvsal1996intrauterine}, which models expected weight as a function of GA and is the established reference in the Scandinavian setting of this cohort; the same reference is used for both estimators. We define the STD interval
\begin{equation}
\mathrm{STD} = \mathrm{GA}_{\mathrm{birth}} - \mathrm{GA}_{\mathrm{scan}} \quad\text{(days)},
\end{equation}
which is computed from GA and is therefore independent of either model. The cohort spans scan GA $34.0\pm3.0$ weeks (range 28--42) and STD $38.8\pm22.2$ days, with maternal age $30.5\pm5.2$ years, BMI $25.0\pm5.9$, and 41\% of triplets at parity ${\leq}1$. Quantile bins used throughout (STD, GA, PS, BMI, age) are equal-count, so each holds ${\approx}10\%$ of triplets for the per-factor gradients and one quarter for the two-factor maps; parity is grouped as ${\leq}1$ versus $\geq2$. All analyses are computed at the pregnancy level, and the DL model's training and validation splits were performed at the pregnancy level in the originating study~\cite{mikolaj2025predicting}, so no woman contributes to both model training and our held-out analysis.

\subsubsection{Models.}
The DL model is a multi-branch convolutional regressor that estimates fetal weight from images together with PS~\cite{mikolaj2025predicting}; we use its predictions on the held-out test set and refer to that study for its architecture, optimizer, loss, and training and validation protocol. The Hadlock formula is a baseline that consumes only three scalar biometric measurements~\cite{hadlock1985estimation}. This contrast is the methodological lever of our study: because Hadlock is a fixed formula to three geometric measurements and performs no representation learning, any error structure it shares with the DL model cannot arise from learned image features, and instead points to an effect intrinsic to the measurements or the prediction target; structure present in the DL model but absent in Hadlock localises a model-specific, representation-driven effect. We evaluate both models by their Mean Relative Error, $\mathrm{MRE}=100\,|\hat{y}-y|/y$, where $\hat{y}$ is the estimated and $y$ the reference weight.

\subsubsection{Stage 1: unsupervised slice discovery.}
To let the data reveal error structure without pre-specifying factors, we cluster the DL penultimate-layer embeddings~\cite{eyuboglu2022domino,olesen2024slicing}. The clustering uses only these embeddings, with no access to error or ground-truth labels, so the discovered slices are not constructed to align with error; this is the sense in which the discovery is unsupervised. We reduce the embeddings by principal component analysis retaining 99\% of variance (16 components) and fit a Gaussian mixture model via expectation-maximisation~\cite{dempster1977maximum}, partitioning examinations into $k$ clusters (``slices''). Rather than commit to a single resolution, we report results across $k\in\{5,8,10,18\}$ and treat agreement across these values as a robustness check. Each slice is characterised by its mean MRE and by its composition across quantile bins of every candidate factor; we report the Spearman rank correlation~\cite{spearman1904proof} between each factor's per-slice mean and per-slice error.

\subsubsection{Stage 2: model-agnostic disentanglement.}
Slice membership derives from the DL embedding and is not a fair partition for the Hadlock formula. For cross-model analysis we instead partition examinations by clinical, acquisition, and demographic attributes, using quantile bins of STD, GA, PS, BMI, maternal age, and parity, and apply identical partitions to both models; we call this joint use of metadata partitions \emph{intersectional metadata partitioning}, meaning subgroups defined by the combination of two or more such attributes rather than one at a time. We report (i) two-factor intersectional MRE maps with a shared colour scale, and (ii) a standardized partial regression $\mathrm{MRE}\sim \mathrm{PS}+\mathrm{GA}+\mathrm{STD}+\mathrm{BMI}+\mathrm{Age}+\mathrm{Parity}$ fit per model, quantifying each factor's contribution with the others held fixed. Coefficients are expressed in MRE percentage points per standard deviation of the predictor, with 95\% confidence intervals. Because every candidate factor enters the same model, each coefficient already adjusts for all the others, so a factor that merely co-varies with STD, or with any other clinical variable, cannot retain a spurious effect; this is what lets us test whether STD is a genuine driver rather than a proxy for a co-occurring clinical profile.

\section{Results}

\subsubsection{High-error subgroups are jointly temporal and acquisition-extreme.}
\begin{figure}[t]
\centering
\includegraphics[width=\textwidth]{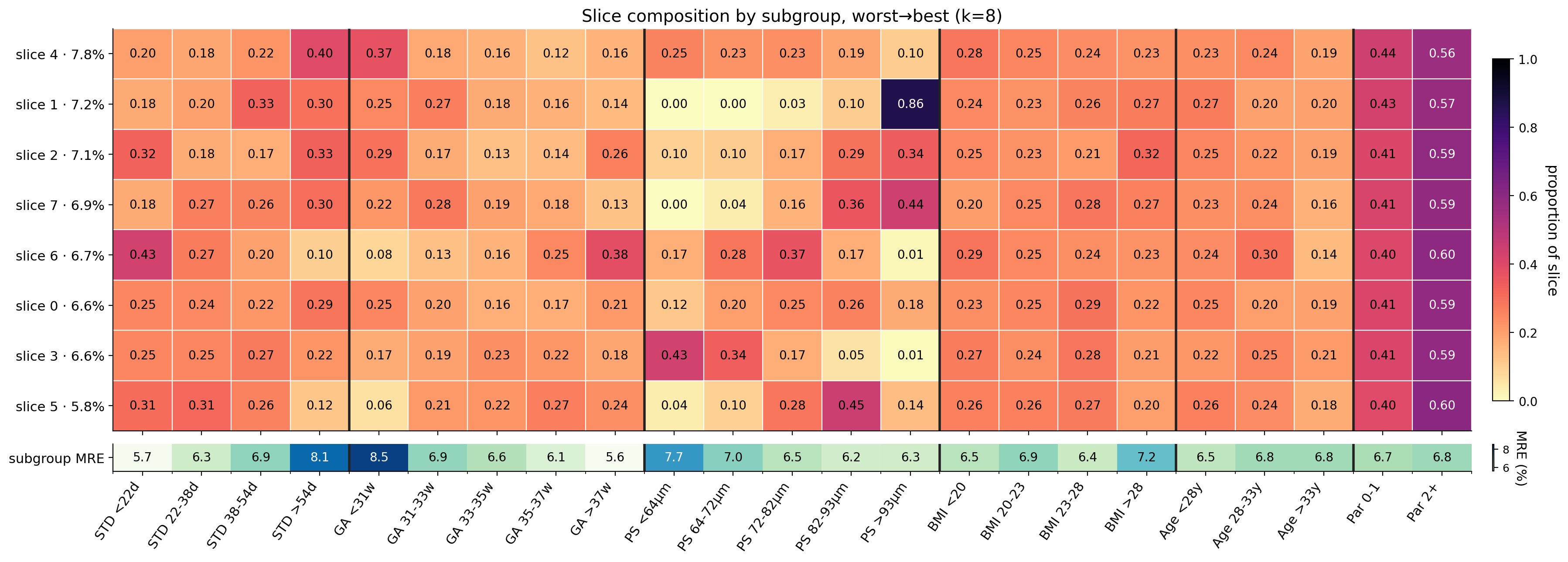}
\caption{Slice composition (Stage 1, $k=8$). Rows are discovered slices ordered by DL error worst$\to$best ; columns are quantile-bin subgroups; cell value is the proportion of the slice in that subgroup. The lower strip shows each subgroup's mean DL MRE.}
\label{fig:leaderboard}
\end{figure}

Fig.~\ref{fig:leaderboard} shows that discovered slices differ substantially in error (5.8--7.8\% MRE at $k=8$), confirming the embedding encodes error-relevant structure. Their composition is entangled: the worst slices are enriched in long STD intervals and early GA. Separately, the embedding also forms near-pure PS slices (one is 86\% highest-PS, another 43\% lowest-PS), but these sit across the error range rather than at the worst end. Across slices, error correlates strongly with the temporal factors GA and STD, and weakly with PS, a pattern stable across $k\in\{5,8,10,18\}$ (GA and STD $|\rho|=0.68$ to $0.90$, significant at every $k$; PS $|\rho|\le0.38$, never significant). Among these, a correlation between temporal features (GA, STD) and error is expected: as our ground truth label is based on birth weight, the weight prediction task is easier at higher GA/lower STD. As such, these temporal factors are the candidates for confounding any apparent acquisition effect. By contrast, PS has no inherent relationship with birth weight, so any relation to error could therefore indicate that some acquisition factors give more informative images than others. To determine whether this reflects a genuine acquisition effect or temporal confounding, we next carried out two complementary analyses to disentangle the effects of PS and STD. 

\begin{figure}[t]
\centering
\includegraphics[width=0.49\textwidth]{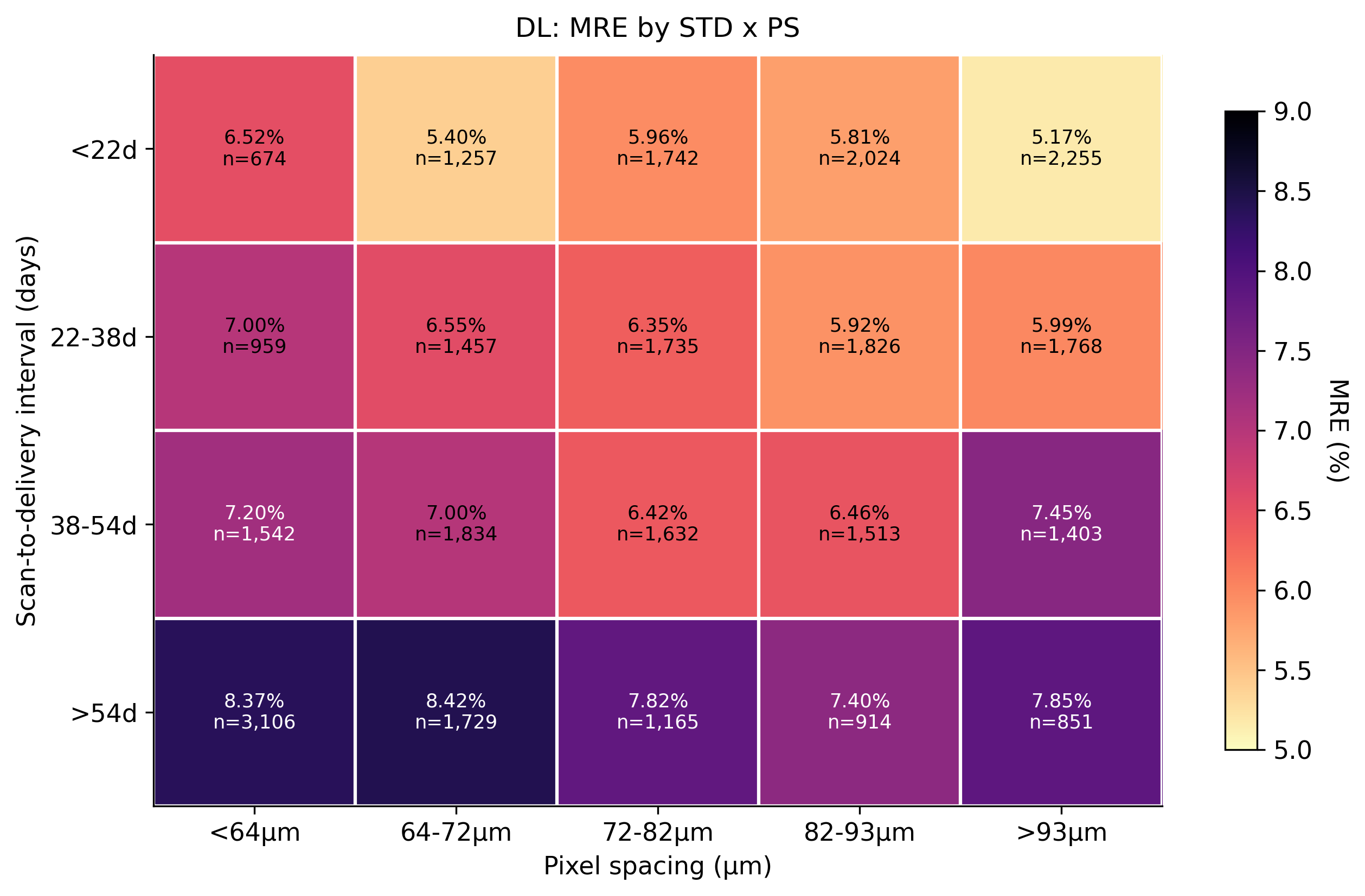}
\hfill
\includegraphics[width=0.49\textwidth]{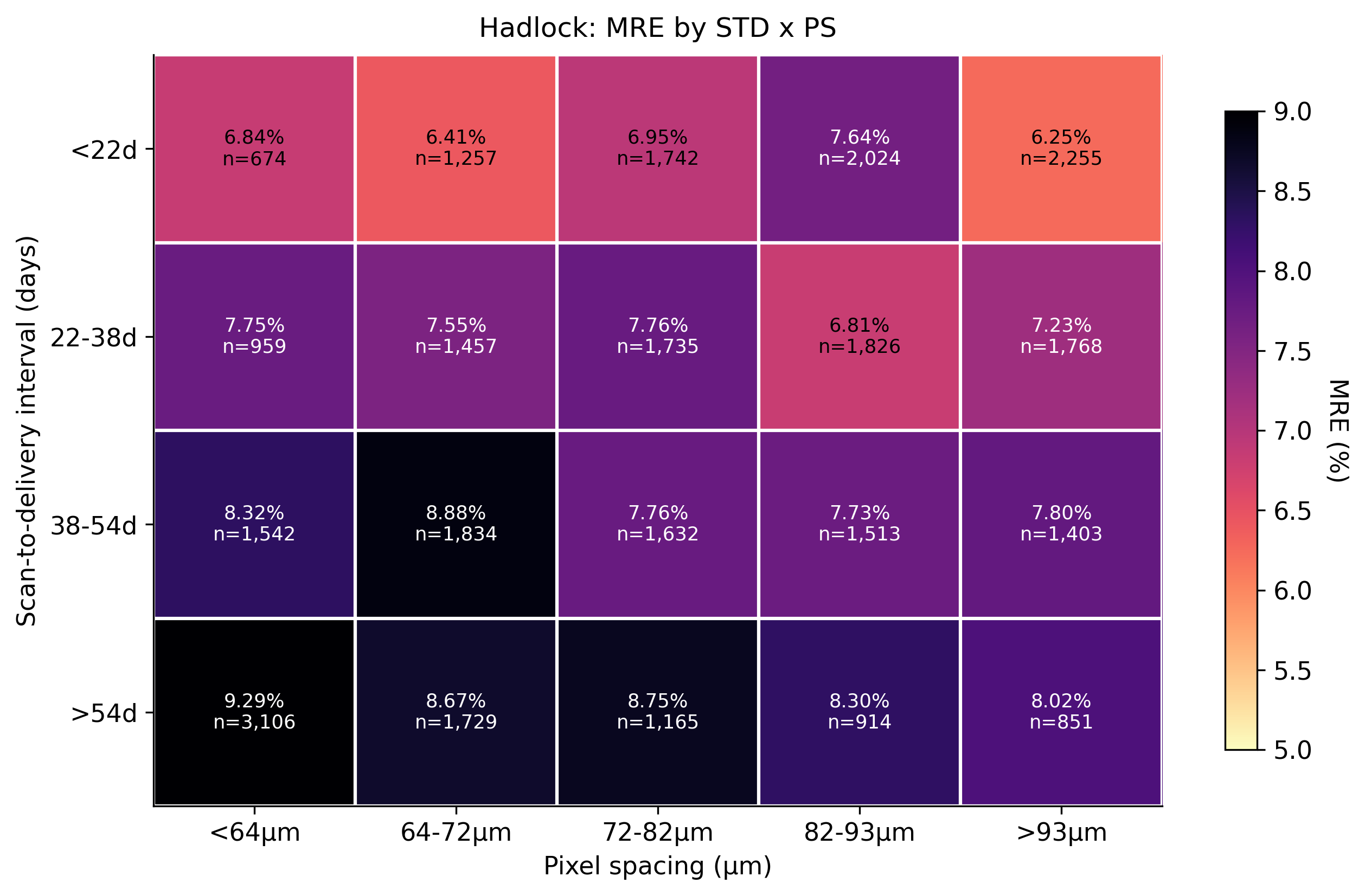}
\caption{Intersectional error maps (Stage 2): STD against PS for the DL model (left) and the Hadlock formula (right). In both, error rises steeply down each column (increasing STD) and changes far less across each row (increasing PS). The same temporal-dominant pattern in Hadlock shows the effect is not a DL artefact.}
\label{fig:heatmap}
\end{figure}

\begin{table}[]
\centering
\caption{Standardized partial-regression coefficients (MRE percentage points per $+1$ SD), $\mathrm{MRE}\sim\mathrm{PS}+\mathrm{GA}+\mathrm{STD}+\mathrm{BMI}+\mathrm{Age}+\mathrm{Parity}$, with 95\% confidence intervals. The temporal factors STD and GA dominate for both models, most for DL; PS is small once time is controlled, and the demographic factors (age, parity) are smaller still. Confidence intervals excluding zero indicate factors whose adjusted contribution is statistically distinguishable from none.}
\label{tab:reg}
\begin{tabular}{lcc}
\toprule
Factor & Hadlock $\beta$ (95\% CI) & DL $\beta$ (95\% CI) \\
\midrule
Scan-to-delivery interval & $+0.37\ (0.21,\,0.54)$ & $+0.56\ (0.41,\,0.70)$ \\
Gestational age & $-0.30\ (-0.46,\,-0.13)$ & $-0.30\ (-0.45,\,-0.15)$ \\
Pixel spacing & $-0.29\ (-0.38,\,-0.20)$ & $-0.17\ (-0.25,\,-0.09)$ \\
Maternal BMI & $+0.14\ (0.06,\,0.22)$ & $+0.11\ (0.04,\,0.19)$ \\
Maternal age & $+0.14\ (0.05,\,0.22)$ & $+0.09\ (0.02,\,0.17)$ \\
Parity & $+0.12\ (0.04,\,0.21)$ & $+0.04\ (-0.03,\,0.12)$ \\
\midrule
Overall MRE & $7.78\%$ & $6.74\%$ \\
\bottomrule
\end{tabular}
\end{table}

\begin{figure}[t]
\centering
\includegraphics[width=0.49\textwidth]{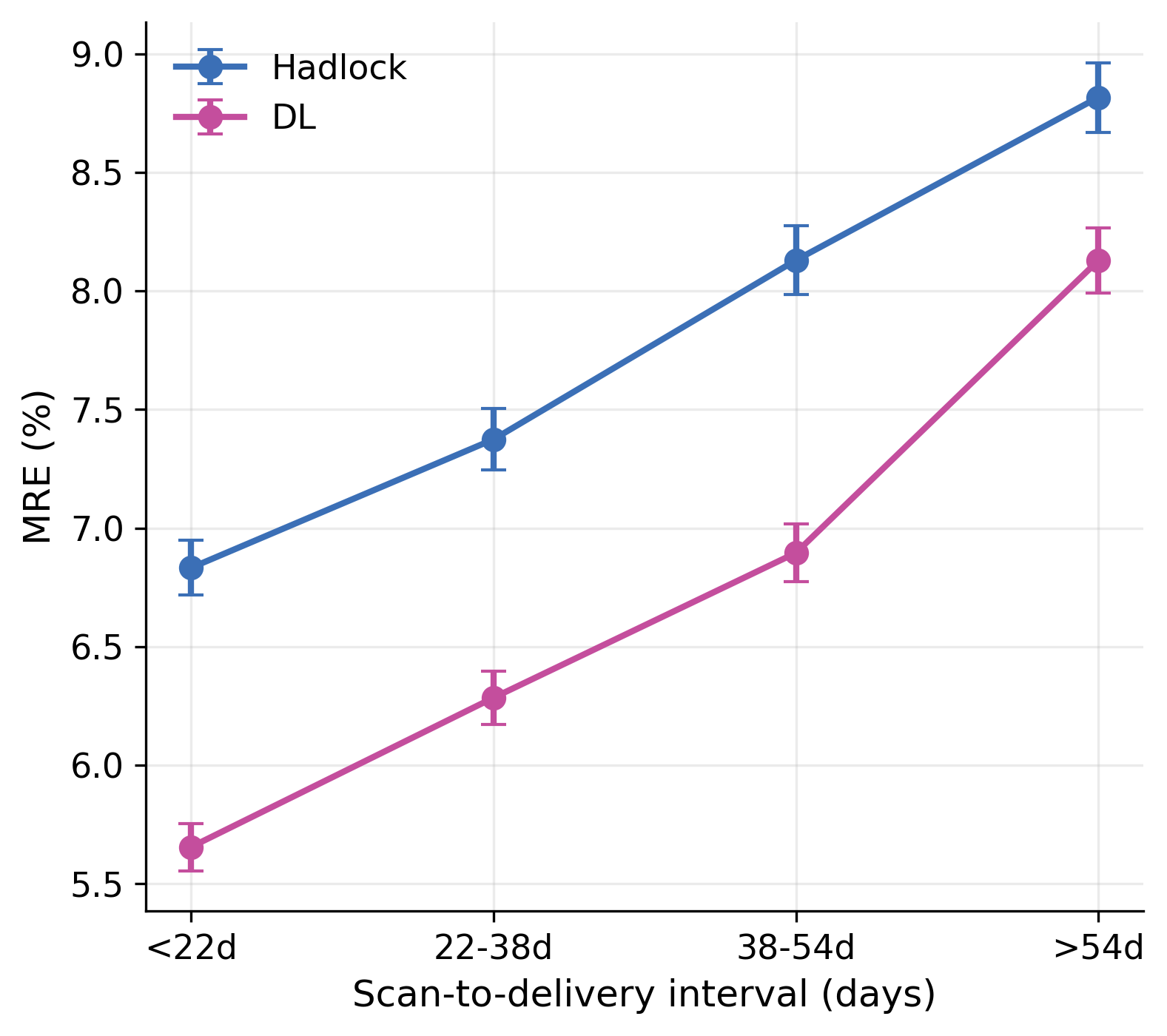}
\hfill
\includegraphics[width=0.49\textwidth]{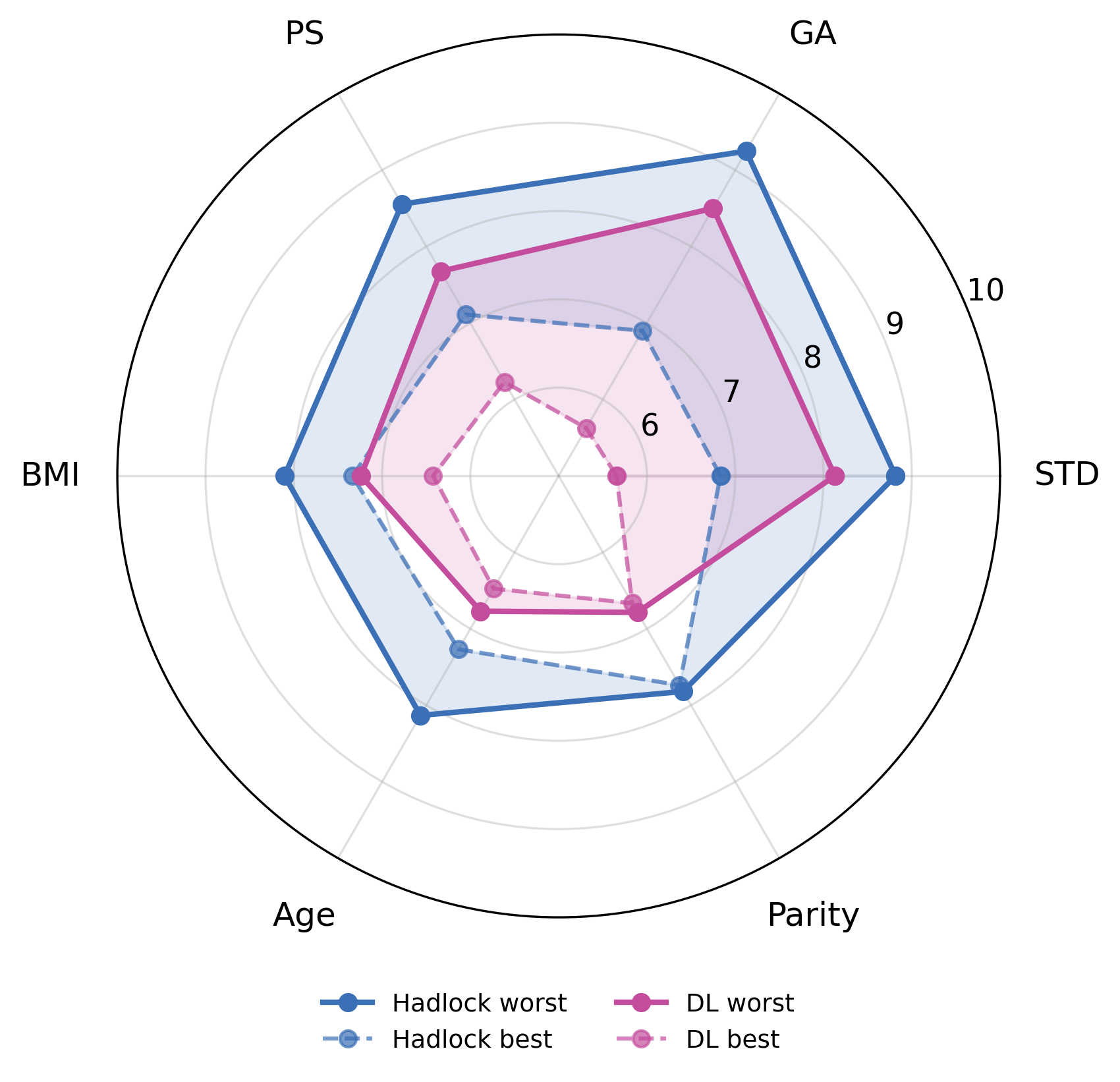}
\caption{Cross-model error structure (Stage 2). \textbf{Left:} mean error rises with the STD for both models; the DL model starts lower and climbs more steeply ($+2.47$ vs.\ $+1.98$ \% across the range), so the gap between them shrinks at long STD. \textbf{Right:} best- versus worst-subgroup error per factor (shaded band = error spread). Both models spread widest on the temporal axes (STD, GA) and narrowest on the demographic ones, with the DL model's band lying inside Hadlock's on every factor. PS shows a sizeable band as well, but this is a \emph{marginal} spread: it largely reflects PS's co-variation with the STD and shrinks after adjustment (Table~\ref{tab:reg}).
}
\label{fig:gradient}
\end{figure}

\subsubsection{Disentanglement: time dominates, pixel spacing (PS) is partially confounded but retains a small residual.} The intersectional maps in Fig.~\ref{fig:heatmap} show that the temporal (STD) and acquisition (PS) factors are not completely disentangled: the diagonal color gradient indicates that they move error together, though unequally. Inspections of single columns and rows show that they also move error independently, and that STD has a higher impact on error than PS: within any fixed STD row, error varies only modestly across PS (by roughly one percentage point), whereas within any fixed PS column it rises steeply with the interval, from $\sim$5.5\% below 22 days to $\sim$8.4\% above 54 days, about three percentage points. STD thus moves error $\sim$3$\times$ more than PS. Similar behavior is observed for the Hadlock formula.

The partial regression (Table~\ref{tab:reg}) confirms this. For the DL model, STD has by far the largest coefficient ($\beta=+0.56$), more than three times the PS magnitude. PS shows a sizeable \emph{marginal} association, including a large best-versus-worst gap (Fig.~\ref{fig:gradient}, right) and a visible heatmap gradient, but this collapses to a small residual once time is controlled ($\beta=-0.17$). We therefore read PS as a small but real effect whose apparent prominence is largely temporal confounding. Because the same pattern appears in both the DL model and the biometry-based Hadlock formula, this residual PS-error association need not reflect optimizable acquisition settings; it could instead reflect unobserved differences in acquisition conditions.

\subsubsection{The temporal effect is intrinsic, and steepest for the DL model.} Both models degrade monotonically as the STD interval grows (Fig.~\ref{fig:gradient}, left). The Hadlock formula degrades as well ($\beta_{\mathrm{STD}}=+0.37$; gradient $+1.98$ percentage points across the interval range), and this behaviour of the baseline is what lets us reason about the underlying mechanism. Two distinct channels can inflate measured error as the interval grows. The first is a \emph{reference} channel that affects any estimator, even a perfect one: the reference label is the birth weight back-calculated to scan time along a population-average growth curve, and individual fetuses deviate from that average by an amount that compounds over a longer scan-to-birth gap, so the target itself becomes noisier at long intervals. The second is a \emph{model} channel specific to the estimator's own predictions: long-interval and correspondingly earlier-gestation examinations are rarer in training and involve smaller anatomy with proportionally noisier measurements, so the learned mapping from image to weight is less reliable in that regime. Because both models are scored against the same reference labels, they share the reference channel. If label noise alone drove the effect, both models would degrade equally with the interval. Instead, the DL model's gradient is markedly steeper ($\beta_{\mathrm{STD}}=+0.56$ vs.\ $+0.37$ for Hadlock, a factor of $\sim$1.5; $+2.47$ vs.\ $+1.98$ percentage points across the interval range), so the steeper climb reflects the DL model's own behaviour at long intervals, not the shared reference label.

The DL model has lower error than Hadlock in every subgroup, including the longest STD ($8.16\%$ vs.\ $8.82\%$ at STD${>}54$d). What varies is the size of its advantage: the DL model leads by ${\sim}1.1$--$1.3$ percentage points at short and medium STD but by only $0.66$ points at the longest, because its error climbs more steeply with the STD. The two estimators thus converge (the gap between them shrinks) as the STD grows, while the DL model stays more accurate throughout. A per-factor analysis (Fig.~\ref{fig:gradient}, right) shows both models spread widest on the temporal axes and narrowest on the demographic ones, with the DL model's band lying inside Hadlock's on every factor.

Adjusting for demographics does not change this picture. Maternal age and parity, the demographic factors examinable in this cohort, carry small coefficients (age $\beta=+0.09$ for DL and $+0.14$ for Hadlock; parity $+0.04$ and $+0.12$), comparable in magnitude to BMI but much smaller than STD. Parity is significant for Hadlock but not for DL. They also fail to organise the high-error slices: their composition is nearly uniform across slices (Fig.~\ref{fig:leaderboard}). Parity in particular illustrates why a single slice-level view is unreliable: it can appear associated with error at some cluster counts, but this association is unstable across $k$ and disappears at larger cluster counts, the hallmark of a small-sample artifact rather than a real driver. Its small adjusted coefficient (Table~\ref{tab:reg}) confirms this. More broadly, because STD retains its large coefficient in the same model that contains GA, BMI, age, and parity, its effect is not a stand-in for a co-occurring clinical profile.

\section{Discussion}
Our central finding is methodological: in fetal weight estimation, an apparent acquisition bias (PS) is largely explained by temporal confounding (STD), revealed only through intersectional disentanglement. Unsupervised slice discovery, an increasingly common first step in bias auditing~\cite{eyuboglu2022domino,olesen2024slicing,oakden2020hidden}, is useful for hypothesis generation, but insufficient to draw these conclusions, because the discovered slices entangle acquisition and time and, read na\"ively, suggest PS bias. This complements work showing that disparities can persist beyond data representation~\cite{dawood2025racial,glocker2023risk}, and that intersectional structure matters for fairness analysis~\cite{buolamwini2018gender,zhang2025intersectional,chen2024fairness}.

The biometric Hadlock baseline robustly supports our conclusion, and helps partially decompose the effect. As Hadlock performs no representation learning yet still degrades with STD, much of the temporal effect is attributable to the shared reference channel rather than to learned image features. Meanwhile, the DL model's additional sensitivity isolates a model-specific component. Practically, weight estimates become less reliable as scan-to-delivery time increases, regardless of acquisition quality, because the effect is largely target-driven. This caveat is invisible both to overall accuracy, where the model is superior, and to standard demographic subgroup analysis.

Our argument is not that acquisition never matters. Maternal BMI, in contrast to PS, retains a small but significant effect after adjustment ($\beta=+0.11$ DL, $+0.14$  Hadlock), consistent with the well-documented degradation of ultrasound image quality at higher maternal BMI~\cite{chung2023obstetric,han2019optimize}. A biological component cannot be excluded, since maternal BMI correlates with fetal size and macrosomia risk. The contrast is the point: our analysis distinguishes a genuine, if modest, acquisition-difficulty factor (BMI) that survives adjustment from one (PS) whose apparent effect is largely temporal confounding, whereas STD dominates both.

\paragraph{Limitations.}
The cohort is single-country and ethnically homogeneous, limiting analysis of ethnicity-related disparities and reflecting the representation imbalance motivating fairness auditing. The reference target is back-calculated, limiting precise separation of reference and model effects; controlled decomposition and selection effects in earlier-scanned high-risk pregnancies remain open questions. Because early-gestation examinations involve smaller anatomy, the manual biometric measurements underlying both the Hadlock estimate and the reference are proportionally noisier there, an operator-related source of error that the present data cannot fully separate from the temporal effect. The temporal gradient remains within training support (27\% second trimester~\cite{mikolaj2025predicting}), so it does not reflect out-of-distribution extrapolation.

\section{Conclusion}

Through a case study of fetal weight estimation we have demonstrated how intersectional disentanglement is essential in bias audits: single-factor slice discovery generates hypotheses about error-driving factors, but can misattribute confounded effects. Here, an apparent acquisition bias (PS) is explained by clinical temporal structure (STD), demonstrating how intersectional analysis can overturn na\"ive single-factor interpretations. Because the reference label itself grows noisier as the scan-to-delivery interval widens, separating this reference uncertainty from genuine model bias is a promising next step for fairness auditing wherever ground-truth labels are reconstructed rather than directly observed.

\subsubsection*{Disclosure of Interests.}
The authors have no competing interests to declare that are relevant to the content of this article.

\bibliographystyle{splncs04}
\bibliography{Paper-0019}
\end{document}